\documentclass{article}

\usepackage{arxiv}

\usepackage[utf8]{inputenc} 
\usepackage[T1]{fontenc}    
\usepackage{hyperref}       
\usepackage{url}            
\usepackage{booktabs}       
\usepackage{amsfonts}       
\usepackage{nicefrac}       
\usepackage{microtype}      
\usepackage{lipsum}
\usepackage{graphicx}
\usepackage{tikz}
\usetikzlibrary{decorations.pathreplacing}
\usepackage{subcaption}
\usepackage{caption}
\graphicspath{ {./images/} }

\title{Estimating City-wide operating mode Distribution of Light-Duty Vehicles: A Neural Network-based Approach}

\author{
 Muhammad Usama \\
  Department of Civil and Environmental Engineering\\
 Northeastern University\\
  Boston, MA, 02115 \\
  \texttt{m.usama@northeastern.edu, engrmusama90@gmail.com} \\
   \And
 Haris N. Koutsopoulos \\
  Department of Civil and Environmental Engineering\\
 Northeastern University\\
  Boston, MA, 02115 \\
  \texttt{h.koutsopoulos@northeastern.edu} \\
  \And
 Zhengbing He \\
  Department of Urban Studies and Planning\\
  Massachusetts Institute of Technology\\
  Cambridge,MA 02139 \\
  \texttt{he.zb@hotmail.com, hezb@mit.edu} \\
   \AND
 Lijiao Wang \\
  Department of Civil and Environmental Engineering\\
  Northeastern University\\
  Boston, MA, 02115 \\
  \texttt{wang.liji@northeastern.edu} \\
}

\begin{document}
\maketitle
\begin{abstract}
Driving cycles are a set of driving conditions and are crucial for the existing emission estimation model to evaluate vehicle performance, fuel efficiency, and emissions, by matching them with average speed to calculate the operating modes, such as braking, idling, and cruising.
While existing emission estimation models, such as the Motor Vehicle Emission Simulator (MOVES), are powerful tools, their reliance on predefined driving cycles can be limiting, as these cycles often do not accurately represent regional driving conditions, making the models less effective for city-wide analyses.
To solve this problem, this paper proposes a modular neural network (NN)-based framework to estimate operating mode distributions bypassing the driving cycle development phase, utilizing macroscopic variables such as speed, flow, and link infrastructure attributes. 
The proposed method is validated using a well-calibrated microsimulation model of Brookline MA, the United States. The results indicate that the proposed framework outperforms the operating mode distribution calculated by MOVES based on default driving cycles, providing a closer match to the actual operating mode distribution derived from trajectory data. Specifically, the proposed model achieves an average RMSE of 0.04 in predicting operating mode distribution, compared to 0.08 for MOVES. The average error in emission estimation across pollutants is 8.57\% for the proposed method, lower than the 32.86\% error for MOVES. In particular, for the estimation of CO$_2$, the proposed method has an error of just 4\%, compared to 35\% for MOVES. The proposed model can be utilized for real-time emissions monitoring by providing rapid and accurate emissions estimates with easily accessible inputs.

\end{abstract}


\section{Introduction}
In 2021, the transportation sector in the U.S. accounted for 67\% of the country's total petroleum consumption, with light-duty vehicles responsible for 63\% of this usage \cite{1}. The extensive use of petroleum contributes significantly to greenhouse gas emissions (GHG), with the transportation sector responsible for 29\% of total U.S. GHG emissions in 2022, making it the largest contributor of direct emissions \cite{2}. Given the adverse environmental, social, and economic impacts of transportation-related emissions, researchers and practitioners have been intensively working to quantify these emissions.

Vehicle emissions are influenced by various factors, including driving style, traffic congestion, traffic control devices, vehicle performance, fuel quality, and ambient operating conditions \cite{3,100}. Therefore, existing emission models have incorporated a range of different variables to better reflect the impact of traffic conditions on emissions \cite{4,101}. The current emission modeling system comprises several models developed to estimate traffic emissions. These models can be broadly categorized into two types: fuel-based and travel-based \cite{5}. Fuel-based models directly use fuel consumption data, which is available from tax records, to estimate GHG based on emission factors expressed in grams per unit of fuel consumed. A notable example of the fuel-based models is the Computer Programme to Calculate Emissions from Road Transport (COPERT), developed by the European Environment Agency (EEA)\cite{6}. Whereas, travel-based emission models combine emission factors for specific regions with travel data to generate emission inventories. These models use emission factors expressed in emissions per unit of driving activity, which can be obtained through dynamometer tests or on-road emission testing.

In the U.S., two primary transportation emission models are currently in use: MOtor Vehicle Emission Simulator (MOVES), developed by the U.S. Environmental Protection Agency (EPA), and EMFAC, developed by the California Air Resources Board (CARB). These models estimate emissions using emission factors expressed as grams of emission per unit of driving activity, primarily based on traditional dynamometer tests of predefined driving cycles. As of July 29, 2024, MOVES4 is the latest version and is used by the EPA for State Implementation Plans (SIPs) and transportation conformity analyses outside California \cite{7}.

Driving cycles, also known as driving schedules, are used in emission models for calculation of emissions, certification, and testing of new vehicles and engines \cite{8}. A driving cycle includes data points representing vehicle velocity at various times, reflecting real-world driving scenarios to assess vehicle performance metrics such as emissions and fuel economy. Driving cycles are categorized into modal and transient types: modal driving cycles involve constant acceleration and speed phases, while transient driving cycles feature frequent and dynamic changes in velocity \cite{9}.

In the U.S., the National Renewable Energy Laboratory (NREL) has advanced driving cycle development with its DRIVE (Drive-Cycle Rapid Investigation, Visualization, and Evaluation) tool \cite{10,11}. This tool utilizes GPS and controller area network (CAN) data to create custom driving cycles based on real-world activity. NREL also offers tools like DriveCAT and the Fleet DNA repository, which provide valuable insights for overcoming technical barriers and enhancing transportation technologies \cite{12,13,14,15,16}.
Much research has been conducted to improve driving cycle accuracy. Zhang et al. analyzed start and idling activities to refine emission estimates in MOVES using the FleetDNA and CE-CERT databases, highlighting the need for specific data collection by fleet type \cite{17,18}. Ivani developed driving cycles for residential refuse trucks in New York using 33 parameters, such as minimum, average, maximum, and standard deviation of speed and acceleration, etc. \cite{19}. Shi et al. employed a chase car method and specified 12 parameters, including average road power \cite{20}, while Kamble et al. used 5 velocity and acceleration parameters \cite{21}. Lai et al. included average road resistance among 10 parameters for city-specific bus driving cycles \cite{22}. Other studies, such as those by Galgamuwa et al., Badusha and Ghosh, and Nesamani and Subramanian, have explored various methods and parameters for driving cycle development, with some suggesting weighted factors for different parameters \cite{23,24,25}. Kondaru et al. introduced weighing factors for different parameters according to their importance to develop real-world driving cycle \cite{9}. However, their study does not propose a proper methodology for determining the weighting factors.


Worldwide, several driving cycles have been established, including the Japanese Cycle (JC08) \cite{26}, Federal Test Procedure (FTP-75) \cite{27}, New European Driving Cycle (NEDC) \cite{28}, and Worldwide harmonized Light duty driving Test Cycle (WLTC) \cite{29}, CARB unified (LA92) cycle \cite{30}, and cycles for cities such as Athens, Melbourne, and Beijing \cite{31}. NEDC includes the urban drive and high-speed motorway drive sub-cycles \cite{9}. WLTC, developed by the United Nations Economic Commission for Europe, is the latest and aims to closely resemble real-world driving scenarios worldwide, though it does not account for regional variations. Country-specific or regional driving cycles can provide more accurate vehicle performance predictions in certain areas \cite{9,10}. A common approach to develop driving cycles involves selecting microtrips that best represent speed-time data traces. Microtrips are segments of data bounded by idle modes. The LA01 cycle uses a Monte Carlo simulation approach and Markov process theory to describe actual driving processes, creating cycles that match target Speed-Acceleration-Frequency Distributions \cite{32}. The quality of developed driving cycles depends on the selection of performance measures and the development method used. Most methodologies employ random or quasi-random methods for selecting microtrips \cite{33}.

Developing representative driving cycles is inherently challenging due to the need to capture the diverse and dynamic nature of real-world driving behaviors. These behaviors vary widely according to geography, traffic conditions, driver habits, and vehicle types. The penetration rate of those data is usually low due to the small number of dedicated vehicles used for data collection.
Most developed local driving cycles suffer from small sample sizes collected from a few vehicles over short periods, making it difficult to represent all driving conditions accurately. 

Studies suggest that standardized driving cycles often fail to distinguish between separate phases of urban roads, rural roads, and motorways, leading to inaccuracies in emissions estimation \cite{31,35}. Additionally, drive cycle development requires several parameters and different studies use varying numbers of parameters with different weights, complicating the process further. There is no standardized method for weighing these parameters, adding another layer of complexity to developing accurate and representative driving cycles. EPA’s MOVES model uses 49 drive cycles to represent all driving conditions and vehicle types. At the network level, emissions are estimated based on the link-averaged speed, which is critical in choosing the driving cycle that matches the closest average speed. Based on selected driving cycles, MOVES calculates speed and Vehicle Specific Power (VSP), assigns operating modes based on speed and VSP ranges, and then computes the fraction of time spent in each mode. The operating mode fractions are adjusted to account for the difference between the link's speed and the driving cycle speed through interpolation. This whole process may take longer running times (e.g. several days) to compute emissions in a traffic network on a citywide scale \cite{102}. However, this approach does not account for link features such as speed limit, lanes and traffic control, etc. Consequently, using default driving cycles to estimate the operating mode distributions can be misleading, and developing local driving cycles remains challenging.

Given the challenges and limitations associated with traditional driving cycles and an activity-based model such as MOVES, there is a need for simpler and efficient models that rely on macroscopic traffic variables and network features. By developing models that relate operating mode distributions to easily accessible data, such as average speed, traffic volume, free flow speed, number of lanes, and intersection types, the emission estimation process can be simplified. Such models would provide more accurate and reliable emissions estimates while being easier to implement and less data-intensive, making them highly valuable for researchers and policymakers. Li et al. (2020) was the first to estimate operating mode distributions directly using macroscopic variables by building models between operating mode distributions and average speed to facilitate emission estimation \cite{36}. They found that arterials and collectors have different operating mode distributions even at the same average speed. However, their model only considers average speed and ignores infrastructure-related features that impact driving patterns.

To fill the gap, this study develops a methodology to directly estimate the city-wide operating mode distributions of traffic links in a traffic network. This estimation, using a Modular Neural Networks (MNN), leverages macroscopic traffic variables and link infrastructure features, providing a more efficient and potentially more accurate approach compared to traditional methods.

Currently, the typical application of MOVES for estimating emissions in an area involves using average link speeds to match default driving cycles, which are then used to calculate operating mode distributions and ultimately determine emissions.

\begin{figure}[h]
    \centering
    \includegraphics[width = 6in, trim=0in 2.65in .1in 1.5in, clip]{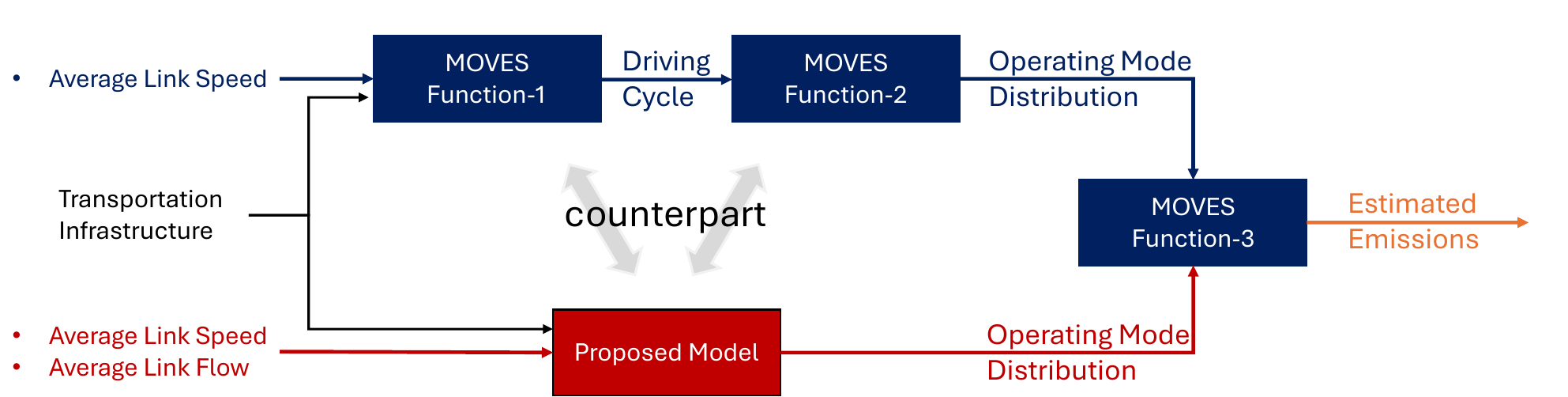}
    \caption{Calculating emissions using MOVES and the proposed model. The proposed model learns traffic dynamics from aggregated traffic and infrastructure data to directly infer accurate operating mode distributions in significantly less runtime. While MOVES uses infrastructure features not related to operating modes and bases its operating modes on drive cycles selected by speed.}
    \label{fig:fig1}
\end{figure}

The proposed approach suggests the use of easily accessible infrastructure and loop detector data to estimate the most appropriate operating mode distributions. The emission estimation process using MOVES and the proposed model are shown in Figure~\ref{fig:fig1}. MOVES Function-1 takes the average speed as input and selects the appropriate driving cycles from its default database. Function-2 processes the selected driving cycles to calculate operating mode distributions through VSP and speed bins. MOVES Function-3 processes the operating mode distribution and all other inputs and calculate the final emissions.


To the best of our knowledge, this is the first study to use a machine learning model to estimate the distribution of all 23 operating modes in MOVES for every link in a city-wide area. The proposed approach has the potential to support traffic emissions analysis for research and decision making at various levels. The contributions of this paper are multifold. First, it develops an MNN designed to directly estimate the operating mode distribution of traffic links leveraging macroscopic traffic variables and link infrastructure features, providing a more efficient and accurate method compared to traditional techniques. Second, it presents a comprehensive framework for estimating traffic emissions on a citywide scale. This framework is designed to operate efficiently within a reasonable runtime without compromising accuracy, thus addressing the limitations of existing methodologies. Third, the model is trained using a well-calibrated traffic simulation, which enhances its robustness and allows for application to other citywide scales without the need for retraining.

In the rest of the paper, we present the proposed approach, followed by a case study to apply and validate the proposed model followed by conclusions with future work recommendations.

\section{Approach}
\label{sec:headings}
The critical step in the traffic emissions estimation process is to infer the appropriate distribution of the operating modes in the given area from the typical traffic and network data. For this, we use MNN to relate traffic characteristics. The approach, in addition to speed, uses other variables such as number of lanes, speed limit, traffic volume road class, and traffic control type, etc. However, and for the purpose of this study, in order to validate the approach we use a microscopic traffic simulation model to obtain detailed trajectories (and hence driving cycles) that allows us to estimate emissions accurately with MOVES and establish the ground truth.

\begin{figure}[htbp]
    \centering
    \includegraphics[width=\linewidth]{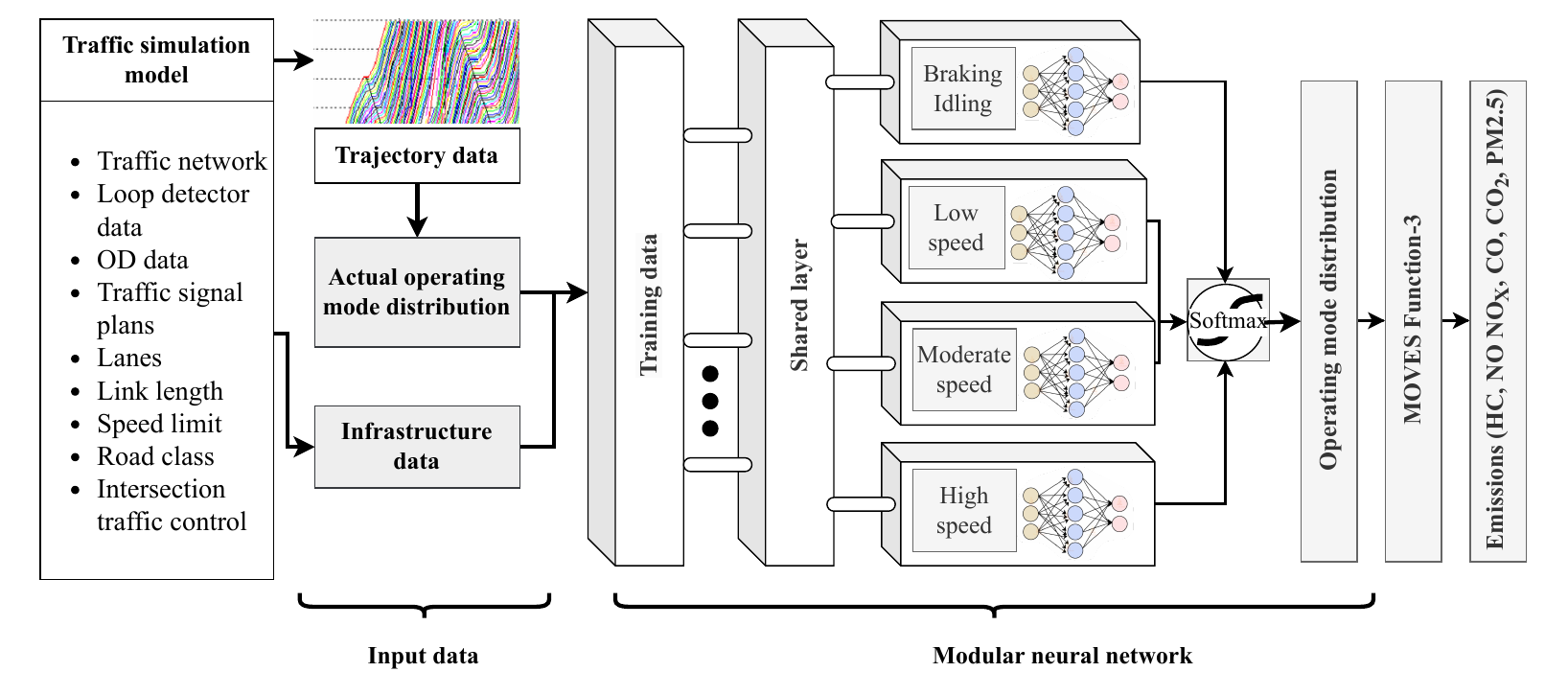}
    \caption{Methodological framework of the proposed model.}
    \label{fig:figure1}
\end{figure}

The methodological framework is shown in Figure~\ref{fig:figure1}. The framework for estimating operating mode distribution involves a traffic simulation model, input data, and an MNN. The traffic simulation model utilizes comprehensive traffic network data such as loop detector data, Origin-Destination (OD) data, traffic signal plans, and various road attributes like lanes, link length, speed limit, road class, and intersection traffic control. The simulation model generates trajectory data, which is processed to create the operating mode distributions. The operating mode distribution, aggregated traffic data and infrastructure features are used as training data for MNN. The simulation data is needed solely for the initial training of the model. Once trained, the model requires only the easily accessible aggregated loop detector and network infrastructure data for subsequent use. MNN network includes a shared layer followed by specialized layers for different speed categories: braking/idling, low speed, moderate speed, and high speed. The outputs from these layers are combined using a softmax function to produce the final operating mode distribution, which are passed to the corresponding MOVES module to estimate the traffic emissions.

A modular neural network architecture is proposed to model the operating mode distribution using infrastructure, average speed, and volume data. MNN employs multiple artificial neural networks (ANNs) as individual modules, each responsible for different aspects of the problem. The system operates by dividing the problem into smaller subproblems and assigning each subproblem to a different module. The results from these modules are then combined to produce the final output of the entire system \cite{37}. MNN are particularly suited for problems with features common to several groups, as they allow for both shared learning and specialized processing \cite{38}.

In the context of MOVES, vehicle operating modes are categorized based on speed and VSP bins. Speed bins group operating modes into major categories such as idling, low speed, moderate speed, and high speed. Each major category, defined by a common speed, is handled by a separate module within the proposed architecture.

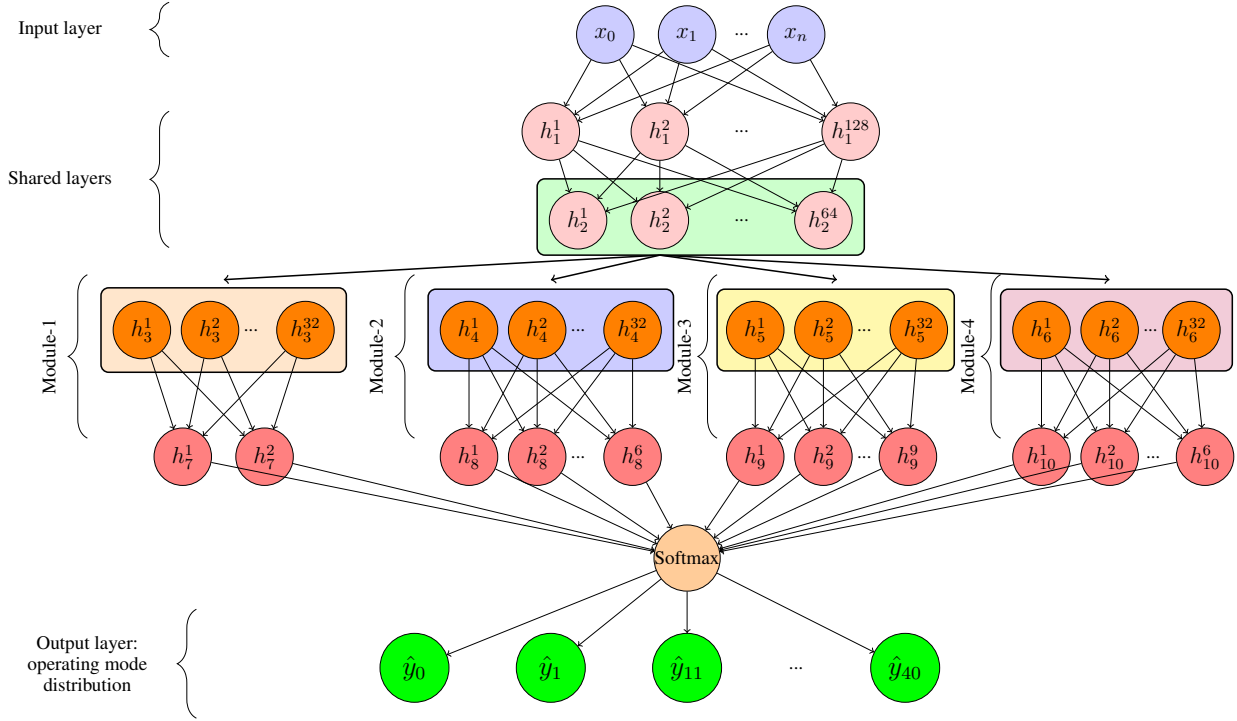
\begin{figure}[htbp]
    \centering
    \resizebox{\textwidth}{!}{
    \begin{tikzpicture}[xscale=-1]
    \begin{scope}[rotate=270] 
        \tikzstyle{unit}=[draw,shape=circle,fill=red!50,minimum size=30pt,inner sep=0pt]
        \tikzstyle{uniti} = [circle, draw, fill=blue!20, minimum size=30pt, inner sep=0pt]
        \tikzstyle{unith} = [circle, draw, fill=red!20, minimum size=30pt, inner sep=0pt]
        \tikzstyle{unitm} = [circle, draw, fill=orange, minimum size=30pt, inner sep=0pt]
        \tikzstyle{unito} = [circle, draw, fill=green, minimum size=30pt, inner sep=0pt]

        \def\xs{1.55} 
        \def\ys{1} 
 
        \node[uniti](x0) at (-1*\xs,3.5*\ys){\large$x_0$};
        \node[uniti](x1) at (-1*\xs,2*\ys){\large$x_1$};
        \node(dots) at (-1*\xs,1) {...};
        \node[uniti](xn) at (-1*\xs,0*\ys){\large$x_n$};
        \draw[draw=black, thick, rounded corners,fill=green!20] 
            (1.1,-1.25) rectangle (2.5,4.75); 
        \node[unith](h11) at (0.15*\xs,4.5){\large$h_1^1$};
        \node[unith](h12) at (0.15*\xs,2.5){\large$h_1^2$};
        \node(dots) at (0.15*\xs,1) {...};
        \node[unith](h1n) at (0.15*\xs,-1){\large$h_1^{128}$};

        \node[unith](h21) at (1.2*\xs,4){\large$h_2^1$};
        \node[unith](h22) at (1.2*\xs,2.5){\large$h_2^2$};
        \node(dots) at (1.2*\xs,1) {...};
        \node[unith](h2n) at (1.2*\xs,-0.5){\large$h_2^{64}$};

        \draw[draw=black, thick, rounded corners, fill=orange!20] 
            (2*\xs,8.25) rectangle (3.0*\xs,12.75);
        \node[unitm](sl1) at (2.5*\xs,12){\large$h_3^1$};
        \node[unitm](sl2) at (2.5*\xs,10.75){\large$h_3^2$};
        \node(dots) at (2.5*\xs,10){...};
        \node[unitm](sln) at (2.5*\xs,9){\large$h_3^{32}$};

        \node[unit](sl1_1) at (4*\xs,11.25){\large$h_7^1$};
        \node[unit](sl1_2) at (4*\xs,9.75){\large$h_7^2$};

        \draw[draw=black, thick, rounded corners, fill = blue!20] 
            (-0.75+2.5*\xs,2.25) rectangle (+0.75+2.5*\xs,6.75); 
        \node[unitm](sl21) at (2.5*\xs,6){\large$h_4^1$};
        \node[unitm](sl22) at (2.5*\xs,4.75){\large$h_4^2$};
        \node(dots) at (2.5*\xs,4){...};
        \node[unitm](sl2n) at (2.5*\xs,3){\large$h_4^{32}$};
        
        \draw[draw=black, thick, rounded corners,fill = yellow!40] 
            (-0.75+2.5*\xs,1.45) rectangle (+0.75+2.5*\xs,-2.9); 
        \node[unit](sl21_1) at (4*\xs,6){\large$h_8^1$};
        \node[unit](sl22_1) at (4*\xs,4.75){\large$h_8^2$};
        \node(dots) at (4*\xs,4){...};
        \node[unit](sl2n_1) at (4*\xs,3){\large$h_8^{6}$};

        \node[unitm](sl31) at (2.5*\xs,0.75){\large$h_5^1$};
        \node[unitm](sl32) at (2.5*\xs,-0.5){\large$h_5^2$};
        \node(dots) at (2.5*\xs,-1.25){...};
        \node[unitm](sl3n) at (2.5*\xs,-2.25){\large$h_5^{32}$};
        
        \draw[draw=black, thick, rounded corners,fill = purple!20] 
            (-0.75+2.5*\xs,-3.75) rectangle (0.75+2.5*\xs,-8); 
        
        \node[unit](sl31_1) at (4*\xs,0.75){\large$h_9^1$};
        \node[unit](sl32_1) at (4*\xs,-0.5){\large$h_9^2$};
        \node(dots) at (4*\xs,-1.25){...};
        \node[unit](sl3n_1) at (4*\xs,-2.05){\large$h_9^{9}$};
        
        \node[unitm](sl41) at (2.5*\xs,-4.5){\large$h_6^1$};
        \node[unitm](sl42) at (2.5*\xs,-5.75){\large$h_6^2$};
        \node(dots) at (2.5*\xs,-6.5){...};
        \node[unitm](sl4n) at (2.5*\xs,-7.25){\large$h_6^{32}$};

        \node[unit](sl41_1) at (4*\xs,-4.5){\large$h_{10}^1$};
        \node[unit](sl42_1) at (4*\xs,-5.75){\large$h_{10}^2$};
        \node(dots) at (4*\xs,-6.55){...};
        \node[unit](sl4n_1) at (4*\xs,-7.5){\large$h_{10}^{6}$};
        
        \node[unit, fill = orange!40](af) at (5.2*\xs,2){Softmax};

        \node[unito, scale=1.2](out1) at (6.5*\xs,7){\large$\hat{y}_{0}$};
        \node[unito, scale=1.2](out2) at (6.5*\xs,4.5){\large$\hat{y}_{1}$};
        \node[unito, scale=1.2](out3) at (6.5*\xs,2){\large$\hat{y}_{11}$};
          \node(dots) at (6.5*\xs,0){...};
        \node[unito, scale=1.2](out4) at (6.5*\xs,-2){\large$\hat{y}_{40}$};

        \draw[->] (af) -- (out1);
        \draw[->] (af) -- (out2);
        \draw[->] (af) -- (out3);
        \draw[->] (af) -- (out4);

        \draw[->] (sl1_2) -- (af);
        \draw[->] (sl1_1) -- (af);
        \draw[->] (sl22_1) -- (af);
        \draw[->] (sl21_1) -- (af);
        \draw[->] (sl2n_1) -- (af);
        \draw[->] (sl31_1) -- (af);
        \draw[->] (sl32_1) -- (af);
        \draw[->] (sl3n_1) -- (af);
        \draw[->] (sl41_1) -- (af);
        \draw[->] (sl42_1) -- (af);
        \draw[->] (sl4n_1) -- (af);

        \draw[->] (x0) -- (h11);
        \draw[->] (x0) -- (h12);
         \draw[->] (x0) -- (h1n);
 
        \draw[->] (x1) -- (h11);
        \draw[->] (x1) -- (h12);
         \draw[->] (x1) -- (h1n);
 
        \draw[->] (xn) -- (h11);
        \draw[->] (xn) -- (h12);
         \draw[->] (xn) -- (h1n);

         \draw[->] (h11) -- (h21);
        \draw[->]  (h11) -- (h22);
         \draw[->] (h11) -- (h2n);
         \draw[->] (h12) -- (h21);
        \draw[->]  (h12) -- (h22);
         \draw[->] (h12) -- (h2n);
         \draw[->] (h1n) -- (h21);
        \draw[->]  (h1n) -- (h22);
         \draw[->] (h1n) -- (h2n);

        \draw[->] (sl1) -- (sl1_1);
        \draw[->] (sl1) -- (sl1_2);

        \draw[->] (sl2) -- (sl1_1);
        \draw[->] (sl2) -- (sl1_2);

        \draw[->] (sln) -- (sl1_1);
        \draw[->] (sln) -- (sl1_2);

        \draw[->] (sl21) -- (sl21_1);
        \draw[->] (sl21) -- (sl22_1);
         \draw[->] (sl21) -- (sl2n_1);
 
        \draw[->] (sl22) -- (sl21_1);
        \draw[->] (sl22) -- (sl22_1);
         \draw[->] (sl22) -- (sl2n_1);
 
        \draw[->] (sl2n) -- (sl21_1);
        \draw[->] (sl2n) -- (sl22_1);
         \draw[->] (sl2n) -- (sl2n_1);

         \draw[->] (sl31) -- (sl31_1);
        \draw[->] (sl31) -- (sl32_1);
         \draw[->] (sl31) -- (sl3n_1);
 
        \draw[->] (sl32) -- (sl31_1);
        \draw[->] (sl32) -- (sl32_1);
         \draw[->] (sl32) -- (sl3n_1);
 
        \draw[->] (sl3n) -- (sl31_1);
        \draw[->] (sl3n) -- (sl32_1);
         \draw[->] (sl3n) -- (sl3n_1);

        \draw[->] (sl41) -- (sl41_1);
        \draw[->] (sl41) -- (sl42_1);
         \draw[->] (sl41) -- (sl4n_1);
 
        \draw[->] (sl42) -- (sl41_1);
        \draw[->] (sl42) -- (sl42_1);
         \draw[->] (sl42) -- (sl4n_1);
 
        \draw[->] (sl4n) -- (sl41_1);
        \draw[->] (sl4n) -- (sl42_1);
         \draw[->] (sl4n) -- (sl4n_1);
         

         \draw[->, thick] 
            (2.5,2.5) -- (2.95,10.5);
        \draw[->, thick] 
            (2.5,2.5) -- (2.955,4.5);
        \draw[->, thick] 
            (2.5,2.5) -- (2.95,-.75);
        \draw[->, thick] 
            (2.5,2.5) -- (2.95,-5.75);
 
        \draw [decorate,decoration={brace,amplitude=10pt,mirror},xshift=-4pt,yshift=0pt] (-2,11.5) -- (-1,11.5) node [black,midway,yshift=+0.cm, xshift = -2cm,rotate=0]{Input layer};

        \draw [decorate,decoration={brace,amplitude=10pt,mirror},xshift=-4pt,yshift=0pt] (0,11.5) -- (2.5,11.5) node [black,midway,yshift=+0.cm, xshift = -2cm,rotate=0]{Shared layers};
        \draw [decorate,decoration={brace,amplitude=10pt,mirror},xshift=-4pt,yshift=0pt] (3,13.0) -- (6,13.0) node [black,midway,xshift=-0.7cm,rotate=90]{Module-1};
        \draw [decorate,decoration={brace,amplitude=10pt,mirror},xshift=-4pt,yshift=0pt] (3,7) -- (6,7.0) node [black,midway,xshift=-0.7cm,rotate=90]{Module-2};
        \draw [decorate,decoration={brace,amplitude=10pt,mirror},xshift=-4pt,yshift=0pt] (3,1.45) -- (6,1.45) node [black,midway,xshift=-0.6cm,rotate=90]{Module-3};
        \draw [decorate,decoration={brace,amplitude=10pt,mirror},xshift=-4pt,yshift=0pt] (3,-3.75) -- (6,-3.75) node [black,midway,xshift=-0.6cm,rotate=90]{Module-4};
        \draw [decorate, decoration={brace, amplitude=10pt,mirror}] 
    (9.0,11) -- (11,11) 
    node [black, midway, yshift=0cm, align=center,  xshift = -2cm] {Output layer: \\ operating mode \\ distribution};
    \end{scope}
    \end{tikzpicture}
    }
    \caption{Modular neural network architecture with input nodes, hidden layers, and output layer.}
    \label{fig:figure2}
\end{figure}

The detailed architecture of MNN is shown in Figure~\ref{fig:figure2}. The architecture includes an input layer, two shared layers, four specialized modules, an integrator, and an output layer. The vector $x = [x_1, x_2, \ldots, x_n]$ represents the input vector and $h_m^k$ represents the $k^{th}$ hidden feature in layer $m$. $\hat{y}_i$ indicates the estimated fraction of operating mode $i$. The input layer passes the inputs to the shared layers. The input vector consists of 13 features, including 6 numeric and 3 categorical features. The categorical features are one-hot encoded, resulting in 13 features. The shared layers are responsible for extracting common features from the input data. The first layer is a fully connected hidden layer with 128 neurons, which processes the input data using the Rectified Linear Unit (ReLU) activation function to capture non-linear relationships. The output from the first layer is then fed into a second fully connected layer with 64 neurons, which continues to refine these features using ReLU. The shared feature extraction serves as a foundation for the specialized modules.

Following the shared layers, the architecture employs four specialized modules to handle different operating modes. For the Braking (0) and Idling (1) modes, the data is processed through a fully connected layer with 32 neurons, followed by another fully connected layer with 2 neurons that outputs the probabilities for bin 0 (B0) and bin 1 (B1). These two modes are important for understanding vehicle behavior in urban traffic conditions where frequent stops and starts are common.

Low speed module handles the prediction of operating modes 11 to 16, which correspond to low-speed conditions. In low-speed module, the data is first processed by a fully connected layer with 32 neurons and then through another fully connected layer with 6 neurons, which predicts the probabilities for bins B11 to B16. Vehicles operating in B11-B16 operating modes are navigating through congested or low-speed areas.

Moderate speed module is responsible for predicting the probabilities of operating modes 21 to 30, by processing the hidden features from the shared layer through a fully connected layer of 32 neurons followed by a layer with 9 neurons. These moderate speeds are characteristic of smoother traffic flow conditions, such as those found on arterial roads. High speed module focuses on high-speed operating modes 33 to 40, which passes hidden features through a fully connected layer with 32 neurons and then a layer with 6 neurons that provides the probabilities for B33 to B40. The outputs from the specialized modules are integrated into a single prediction vector, which is then normalized using a softmax function to produce a probability distribution across all operating modes.

\section{Case Study}
\label{sec:others}
 To validate the proposed model, the paper uses Brookline, Massachusetts, as a case study. Brookline presents a diverse mix of urban and residential traffic patterns. Its road network includes a variety of street types, from major arteries to quiet residential roads. The town features a blend of dense commercial areas, residential neighborhoods and university campuses. This land use mix results in a wide range of traffic scenarios with varying vehicle speeds. Brookline's geographical and traffic characteristics make it a good candidate to apply and validate the proposed model.

\subsection{Traffic Simulation}

 A comprehensive microsimulation model has been developed in TransModeler \cite{39} for the city of Brookline, Massachusetts. The road geometry and lane information are obtained from OpenStreetMap (OSM) \cite{40}. Detailed information, such as lane width, the number of lanes at intersections, and lane connectivity are acquired from Google Street View (GSV) \cite{41}. The study area is divided into 27 traffic analysis zones using MassDOT transportation planning data. Origin and destination were defined at each zone, as well as entry and exist points of the various roads that cross the boundaries for a total of 169 origins. Additionally, traffic signal plans of 57 intersections and fixed loop detector data are obtained from the City of Brookline.

\begin{figure}[htbp]
    \centering
    \includegraphics[width=\linewidth]{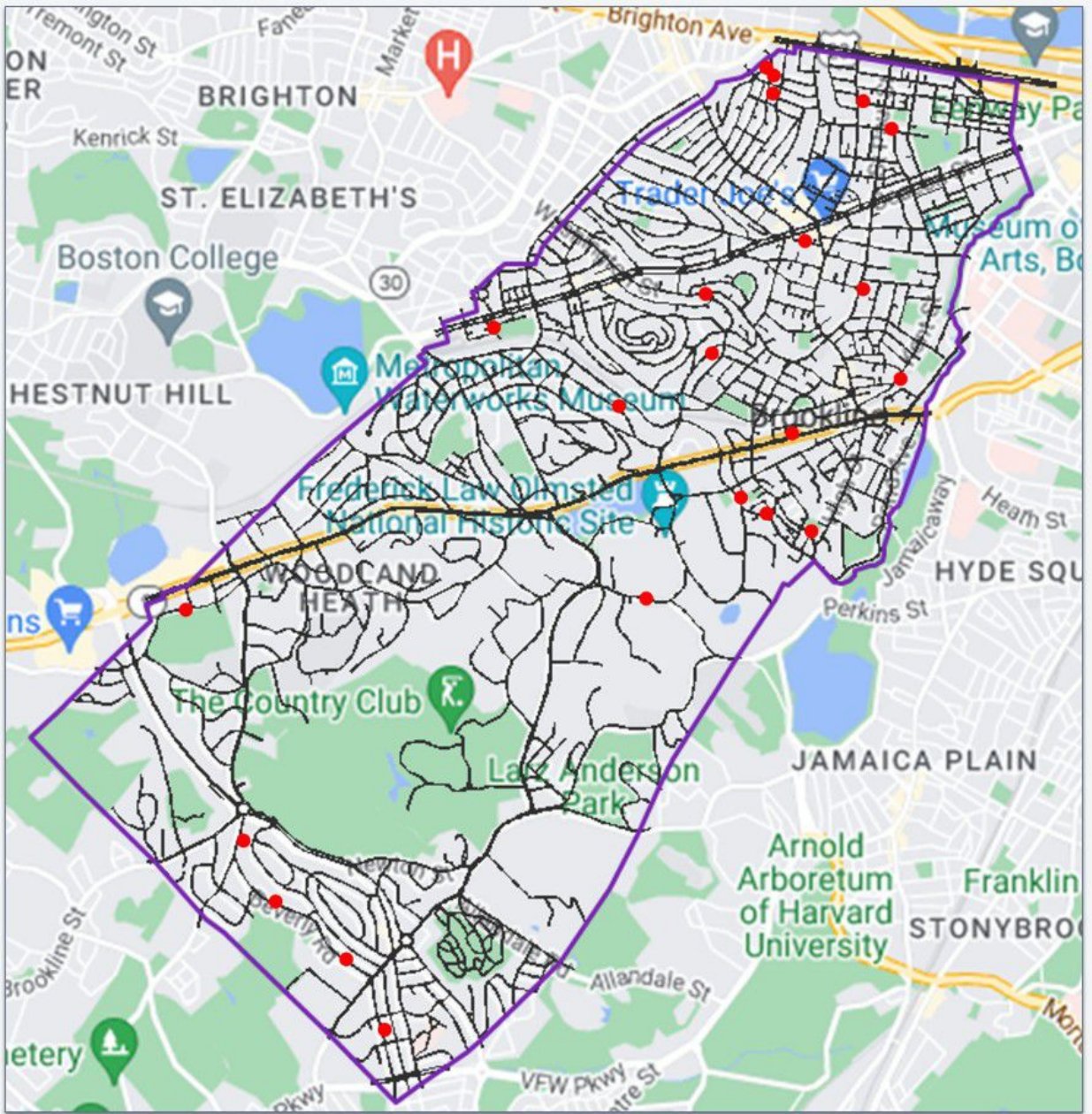}
    \caption{The traffic microsimulation model of the City of Brookline, MA in Transmodeler (The red dots show the location of traffic count sensors).}
    \label{fig:figure3}
\end{figure}

The network comprises 1814 links, each of which may consist of multiple segments to represent changes in road geometry accurately. Figure~\ref{fig:figure3} displays the traffic simulation network of Brookline in TransModeler, highlighting the inclusion of almost every significant link. Red circles mark the locations where traffic count sensors are installed for calibration purposes. The simulation utilizes traffic input data from the morning peak hour (8 AM to 9 AM) to reflect the city's traffic conditions during a critical period accurately.

\subsection{Model Calibration}

The calibration process involves adjusting the model parameters to align simulated traffic patterns with observed data, particularly in the context of OD matrix estimation. Due to its importance, OD matrix estimation has been extensively studied by various researchers. Osorio studied dynamic OD matrix calibration for large-scale networks using simulation-based optimization \cite{42}. Tympakianaki et al. proposed a robust simultaneous perturbation stochastic approximation algorithm for dynamic OD matrix estimation \cite{43,44}. Toledo et al. presented methods for calibration and validation of microscopic models \cite{45,46}. Antoniou et al. presented calibration models and approaches for offline and online dynamic traffic assignment systems \cite{47,48,49}.
\begin{figure}[h]
    \centering
    \includegraphics[width=5in]{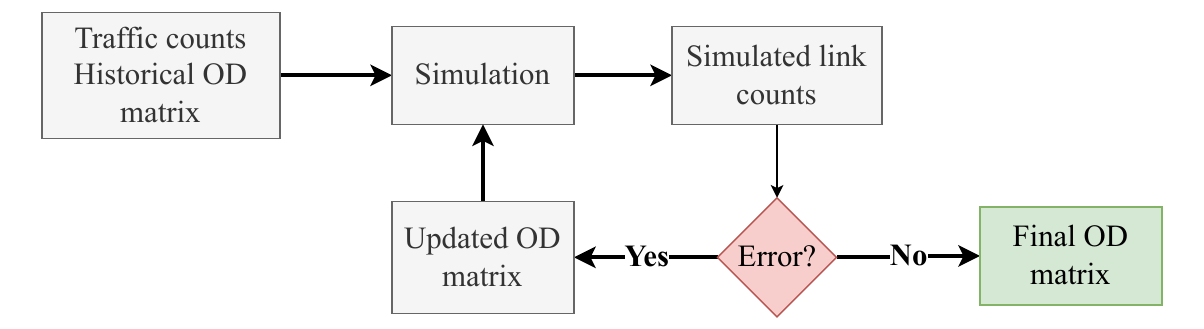}
    \caption{Traffic simulation model Origin-Destination flow calibration.}
    \label{fig:figure4}
\end{figure}

The calibration process is illustrated in Figure~\ref{fig:figure4}. A historical OD matrix (available from previous studies and planning models) is fed into the model, and the simulated traffic counts are then compared against the actual traffic counts obtained from loop detectors placed throughout the network. If the discrepancy between the simulated and actual traffic counts exceeds a predefined error threshold, the OD flows are updated accordingly, and the process is repeated. This iterative process continues until all simulated counts match the actual loop detector counts within an acceptable error threshold.

\begin{figure}[htbp]
    \centering
    \includegraphics[width=5in]{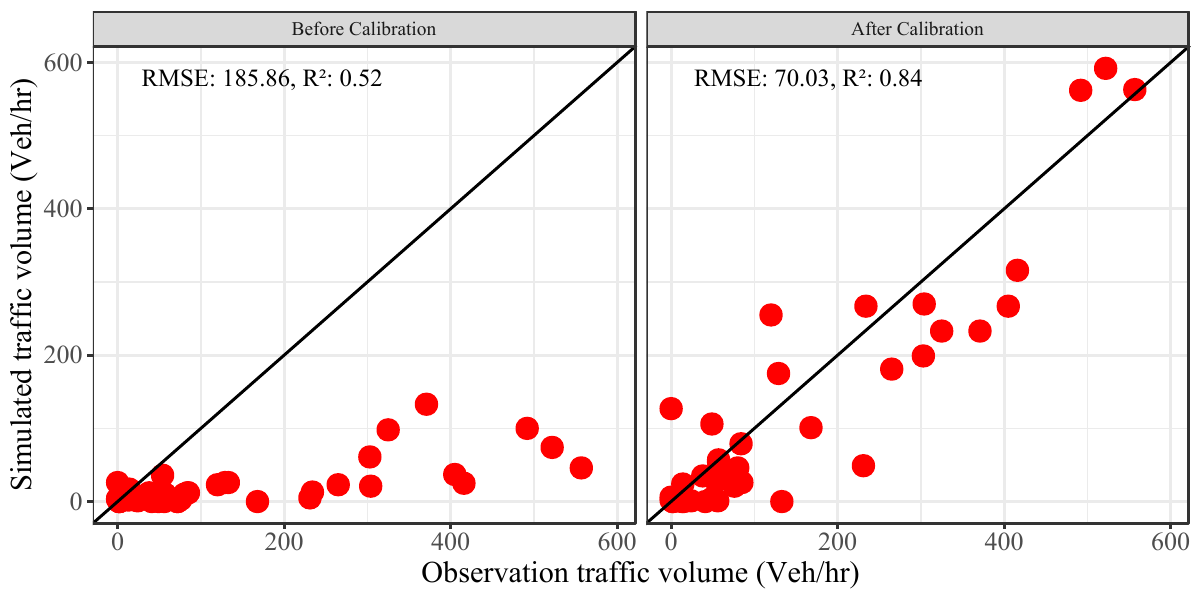}
    \caption{Observed vs simulated traffic counts before and after the calibration.}
    \label{fig:figure5}
\end{figure}

The calibration process ensures that the model accurately reflects the current traffic conditions. Figure~\ref{fig:figure5} shows the observed and simulated traffic counts at the sensor locations before and after the calibration. While, some discrepancies between simulated and observed data still exists, the remaining differences are acceptable for this specific application.

\subsection{Trajectory Data}
Microscopic traffic simulation models output trajectory data for all vehicles during the entire simulation period at a high frequency (e.g. 1Hz). Speed and acceleration are particularly important variables used to calculate VSP, a key input for determining the operating modes of vehicles at a detailed level. VSP is a measure of the power required by a vehicle to overcome various forces such as rolling resistance, aerodynamic drag, and inertia. By calculating VSP, vehicles can be classified into different operating modes, which are indicative of their driving behavior and energy consumption.

\subsection{Operating Mode Distribution}
Table~\ref{tab:VSPBins} presents the operating mode bins as defined by the MOVES model, which categorizes vehicle operating modes based on VSP and speed. For instance, Bin 11 represents conditions where VSP is less than 0 and speed is between 1 mph and 25 mph, while Bin 35 represents conditions where VSP is between 6 and 12, and speed is greater than 50 mph. The braking Bin 0 refers to the condition when instantaneous deceleration is less than 2 mi/hr.sec, or the deceleration of continuous three-second data is less than 1 mi/hr.sec.

The instantaneous VSP at time $t$ is given by:

\begin{equation}
VSP_t = \frac{c_1}{c_2} \frac{Av_t}{m} + \frac{c_1^2}{c_2} \frac{Bv_t^2}{m} + \frac{c_1^3}{c_2} \frac{Cv_t^3}{m} + c_1^2 v_t a_t
\end{equation}

\vspace{10pt}where $A$ is the rolling resistance coefficient ($kW·sec/m$); $B$ is the rotational resistance coefficient ($kW·sec^2/m^2$); $C$ is the aerodynamic drag coefficient ($kW·sec^3/m^3$); $m$ is the vehicle mass (lb); $v_t$ is the instantaneous velocity at time $t$ ($mi/hr$); $a_t$ is the instantaneous acceleration at time $t$ ($mi/hr/sec$); $c_1$ is the conversion factor for speed, and $c_2$ is the conversion factor for vehicle weight.

The trajectory data is processed to calculate VSP for each time step, and operating modes are assigned based on the computed VSP values. The result is a detailed distribution of operating modes across all traffic segments. 

\begin{table}[htbp]
    \caption{EPA MOVES operating mode bins.}\label{tab:VSPBins}
    \begin{center}
        \begin{tabular}{l l l l}
            VSP & 1 mph $\leq$ Speed < 25 mph & 25 mph $\leq$ Speed < 50 mph & 50 mph $\leq$ Speed \\\hline
            0 > VSP & Bin 11 & Bin 21 & Bin 33 \\
            0 $\leq$ VSP < 3 & Bin 12 & Bin 22 &  \\
            3 $\leq$ VSP < 6 & Bin 13 & Bin 23 &  \\
            6 $\leq$ VSP < 9 & Bin 14 & Bin 24 & Bin 35 \\
            9 $\leq$ VSP < 12 & Bin 15 & Bin 25 &  \\
            12 $\leq$ VSP < 18 & Bin 16 & Bin 27 & Bin 37 \\
            18 $\leq$ VSP < 24 &  & Bin 28 & Bin 38 \\
            24 $\leq$ VSP < 30 &  & Bin 29 & Bin 39 \\
            30 $\leq$ VSP &  & Bin 30 & Bin 40 \\
            \hline
            Braking & Bin 0 &  &  \\
            Idling & Bin 1 &  &  \\
            \hline
        \end{tabular}
    \end{center}
\end{table}

\subsection{Infrastructure Data}
The infrastructure data provides a detailed representation of various parameters essential for traffic modeling and analysis. The road segments are characterized by multiple attributes, including the number of lanes, segment length, travel lanes, free flow speed, speed limit, road class, control type, and priority. These attributes collectively define the physical and regulatory environment of the segments, influencing vehicle behavior, route choices, and overall traffic dynamics.

The classification of roads, indicated by the class attribute, categorizes the segments into various types, including arterial, collector, access road, and local street. The control attribute defines the type of traffic control device present at the intersection of the segment's end. A value of 0 indicates that there is no control device, and the segment continues onto another link. Other control types include actuated, pretimed, and roundabout, each representing different traffic signal operations or roundabout presence. Actuated signals adjust their phases based on real-time traffic conditions, whereas pretimed signals follow a fixed schedule. Priority codes are used to determine right-of-way between conflicting turning movements at intersections without explicit signals or signs. 

\subsection{Training}
The simulation model was used to generate a training dataset. The resulting dataset was subsequently divided into training and testing subsets in an 80:20 ratio. 
The features within the data were normalized using the \verb1StandardScaler1 from the \verb1sklearn1 library in python, which standardizes features with zero mean and unit variance \cite{scikit-learn}. To facilitate the training process, the training and testing datasets were converted into \verb1TensorDataset1 objects, and data loaders were employed to handle the data in mini batches \cite{paszke2019pytorch}. Specifically, a batch size of 32 was utilized, meaning that 32 samples were processed before the model parameters were updated. The data loader ensures that during each epoch, the training data is shuffled, thereby promoting better generalization by preventing the model from learning the order of the samples.

\begin{figure}[t]
    \centering
    \includegraphics[width=3.5in]{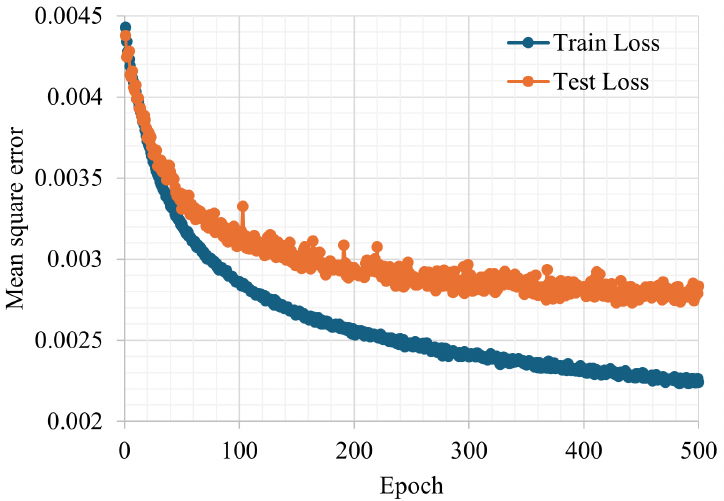}
    \caption{Training and test loss of the proposed model.}
    \label{fig:figure6}
\end{figure}

The model training used the mean square error (MSE) for the loss function, which is a common metric for regression tasks. An Adam optimizer \cite{kingma2014adam} was employed for model optimization, with a learning rate set to 0.001. Adam optimizer is efficient and effective in training deep learning models. The model was trained for 500 epochs. Figure~\ref{fig:figure6} shows the training and test loss curves for the proposed model using MSE as a loss metric. Both curves exhibit a sharp decline in loss during the initial 200 epochs, indicating effective learning. However, beyond 200 epochs, the rate of decrease becomes less pronounced, with the training loss continuing to decrease steadily while the test loss shows only minimal improvement.

\subsection{Evaluation}

The study compares emissions estimated using three distinct approaches to demonstrate the performance of the proposed model. First, ground truth emissions are established by using a microscopic traffic simulation model to obtain detailed vehicle trajectories. These trajectories are then processed to calculate the operating mode distributions, which are subsequently used to estimate emissions accurately using MOVES. Second, MOVES default methodology is utilized to estimate emissions, where link speeds are fed to MOVES and MOVES's default approach is used to select driving cycles and operating mode distributions. Third, the proposed trained model is used to estimate operating mode distributions from traffic and infrastructure characteristics that are then used to estimate emissions.

To evaluate the model performance, we use the root mean square error (RMSE)  and R² score, from python’s \verb1sklearn1 library, of the results using the proposed MNN model and MOVES, in comparison with the ground truth emissions. The results show that the proposed model generally outperforms MOVES, particularly in bins with higher fractions of specific driving conditions.

\begin{figure}[htbp]
    \centering
    \begin{subfigure}[b]{0.48\linewidth}
        \centering
        \includegraphics[width=\textwidth]{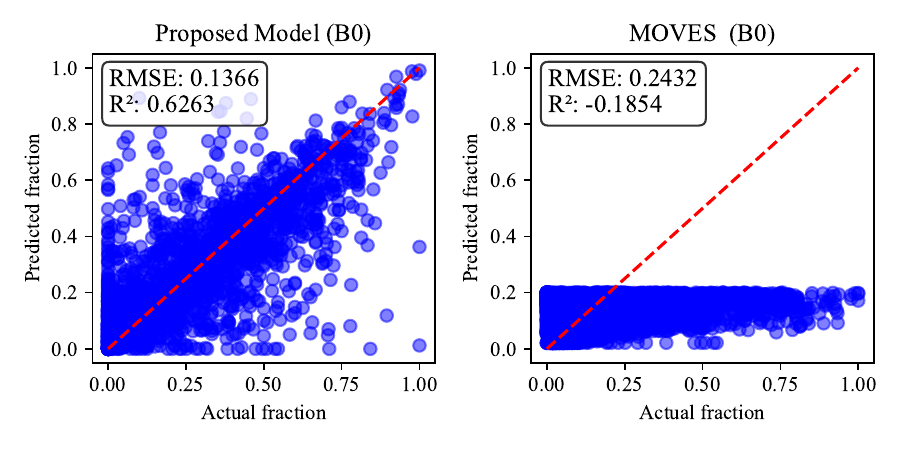}
        \caption{\fontsize{8}{12}\selectfont Estimated vs true operating mode fractions of Bin 0}
        \label{fig:7(a)}
    \end{subfigure}
    \hfill
    \begin{subfigure}[b]{0.48\textwidth}
        \centering
        \includegraphics[width=\textwidth]{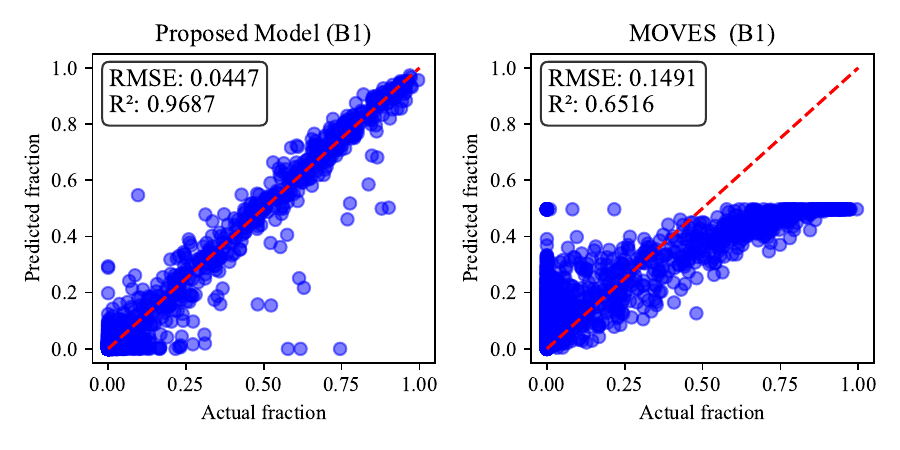}
        \caption{\fontsize{8}{12}\selectfont Estimated vs true operating mode fractions of Bin 1}
        \label{fig:7(b)}
    \end{subfigure}
    \\
    \begin{subfigure}[b]{0.48\textwidth}
        \centering
        \includegraphics[width=\textwidth]{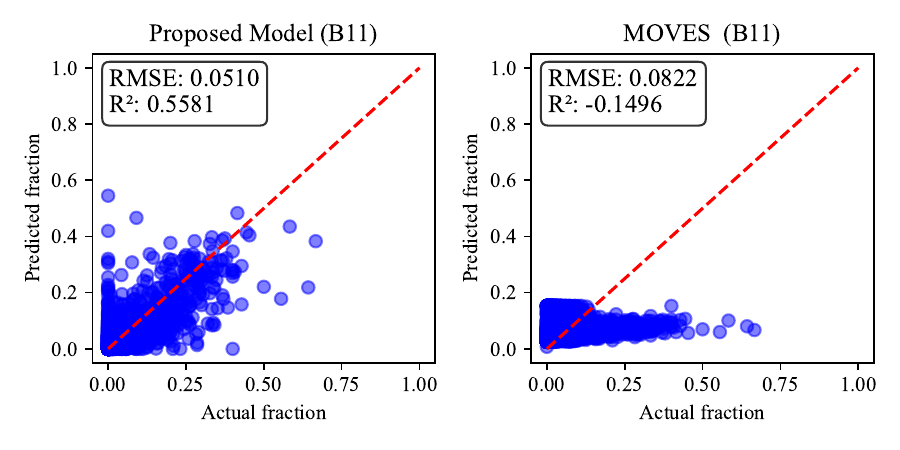}
        \caption{\fontsize{8}{12}\selectfont Estimated vs true operating mode fractions of Bin 11}
        \label{fig:7(c)}
    \end{subfigure}
    \hfill
    \begin{subfigure}[b]{0.48\textwidth}
        \centering
        \includegraphics[width=\textwidth]{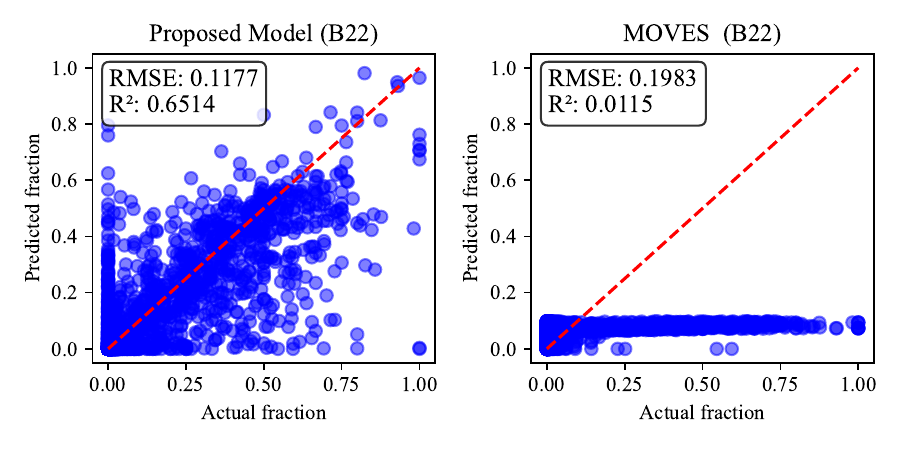}
        \caption{\fontsize{8}{12}\selectfont Estimated vs true operating mode fractions of Bin 22}
        \label{fig:7(d)}
    \end{subfigure}
    \\
    \begin{subfigure}[b]{0.48\textwidth}
        \centering
        \includegraphics[width=\textwidth]{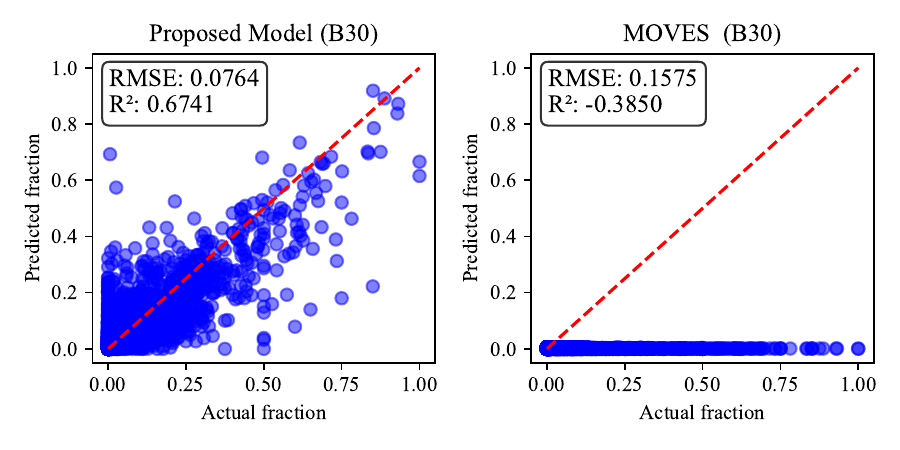}
        \caption{\fontsize{8}{12}\selectfont Estimated vs true operating mode fractions of Bin 30}
        \label{fig:7(e)}
    \end{subfigure}
    \hfill
    \begin{subfigure}[b]{0.48\textwidth}
        \centering
        \includegraphics[width=\textwidth]{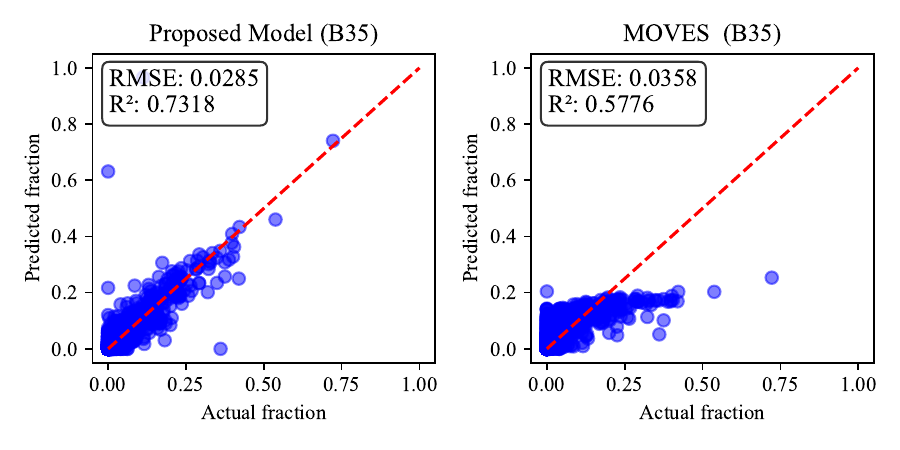}
        \caption{\fontsize{8}{12}\selectfont Estimated vs true operating mode fractions of Bin 35}
        \label{fig:7(f)}
    \end{subfigure}
    \caption{Comparison of operating model fractions obtained using the proposed model and MOVES against ground truth.}
    \label{fig:figure7}
\end{figure}

Figure~\ref{fig:figure7} shows the comparison of operating mode fractions using the proposed model and MOVES. In the figure, bins are selected from each of the four main modules of the proposed model to show their individual training and overall performance in the modular neural network. Each point in the plot represents the actual versus predicted operating mode fractions for a link. Figures~\ref{fig:7(a)} and ~\ref{fig:7(b)} present the estimated operating mode fraction against the actual fraction for Bin 0 and Bin 1 (braking/idling module), respectively. Figures~\ref{fig:7(c)} and ~\ref{fig:7(d)} show the estimated vs. actual fractions for Bin 11 (low-speed module) and Bin 22 (moderate-speed module). Similarly, Figures~\ref{fig:7(e)} and ~\ref{fig:7(f)} present the estimated vs. actual fractions for Bin 30 and Bin 35 (high-speed module), respectively. The results show that bins with higher fractions result in better performance of the proposed model, as evidenced by the closer alignment of the estimated fractions with the actual data.

The best performance is observed for operating mode B1, which represents stop-and-go traffic typically encountered during urban peak hours. It has the lowest RMSE (0.0447) and the highest R² score (0.9687) for the proposed model, indicating an accurate fit to the actual data. The default process in MOVES, while still reasonable in this bin, has higher RMSE (0.1491) and a lower R² score (0.6516). The results indicate that the proposed model is able to capture the dynamics of stop-and-go traffic.

For bins such as B0, B11, and B22, the proposed model also shows better performance with lower RMSE values and higher R² scores compared to MOVES. Interestingly, some bins such as B13, B21, B24, and B30 exhibit negative R² scores for MOVES, which indicate that its predictions are worse than simply using the mean of the actual data \cite{scikit-learn}. It demonstrates that MOVES , which relies on default driving cycles, fails to accurately represent the actual driving conditions for most of the operating modes. In contrast, the proposed model consistently shows better performance, though it is not always perfect for bins having low fractions.

\begin{figure}[htbp]
    \centering
    \includegraphics[width=\linewidth]{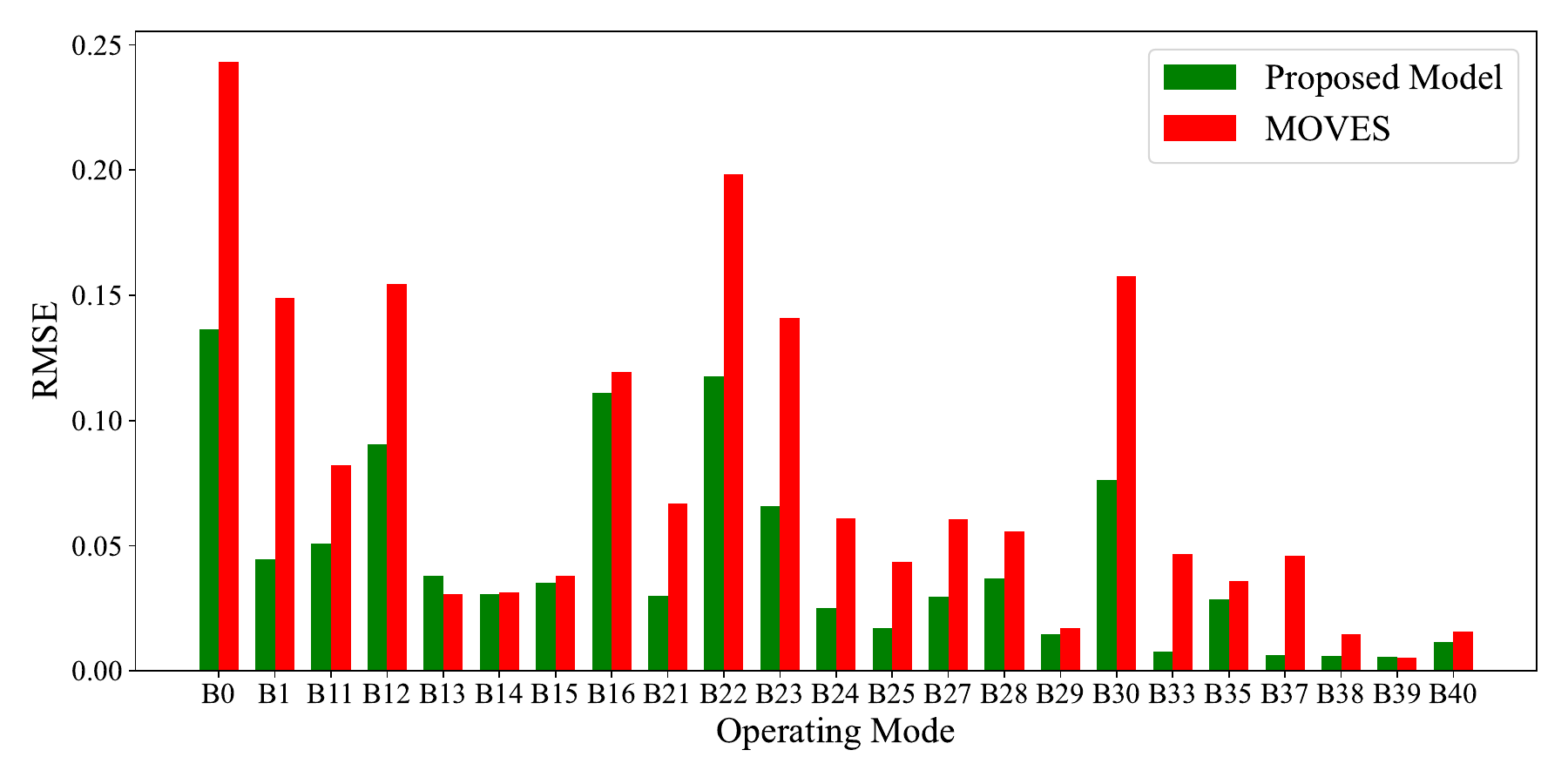}
    \caption{Root mean square error in estimated operating mode fraction using the proposed model and MOVES.}
    \label{fig:figure8}
\end{figure}

Figure~\ref{fig:figure8} illustrates the RMSE in the estimated fraction for each operating mode bin using the proposed model and MOVES. The green and red bars represent the RMSE for the proposed model and MOVES, respectively. The results show that the proposed model consistently exhibits lower RMSE values across nearly all bins compared to MOVES.

\begin{figure}[htbp]
    \centering
    \includegraphics[width=\linewidth]{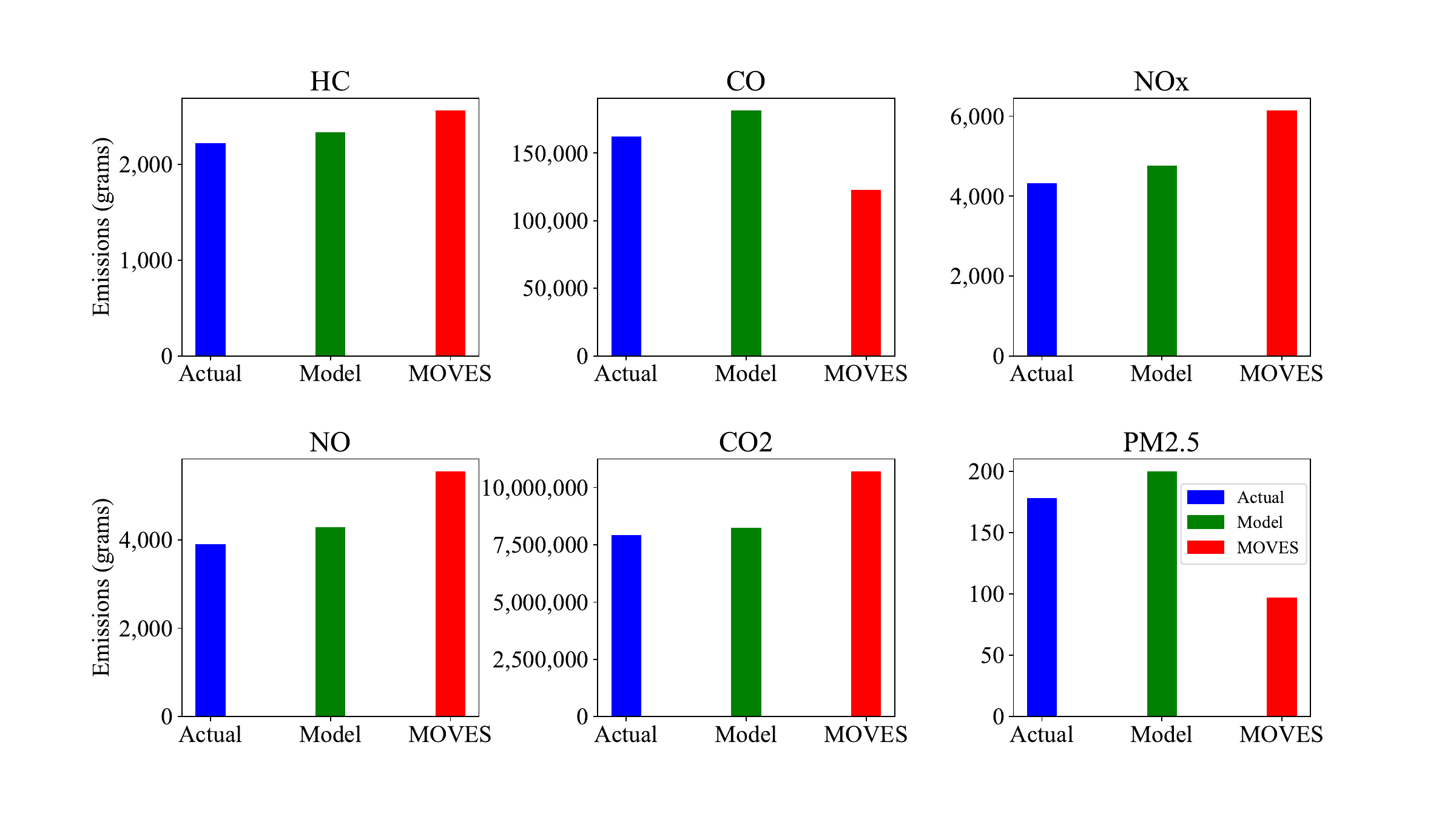}
    \caption{Comparison of pollutant emissions (grams) from the actual trajectory data, the proposed model and MOVES.}
    \label{fig:figure9}
\end{figure}

\begin{figure}[htbp]
    \centering
    \includegraphics[width=5in]{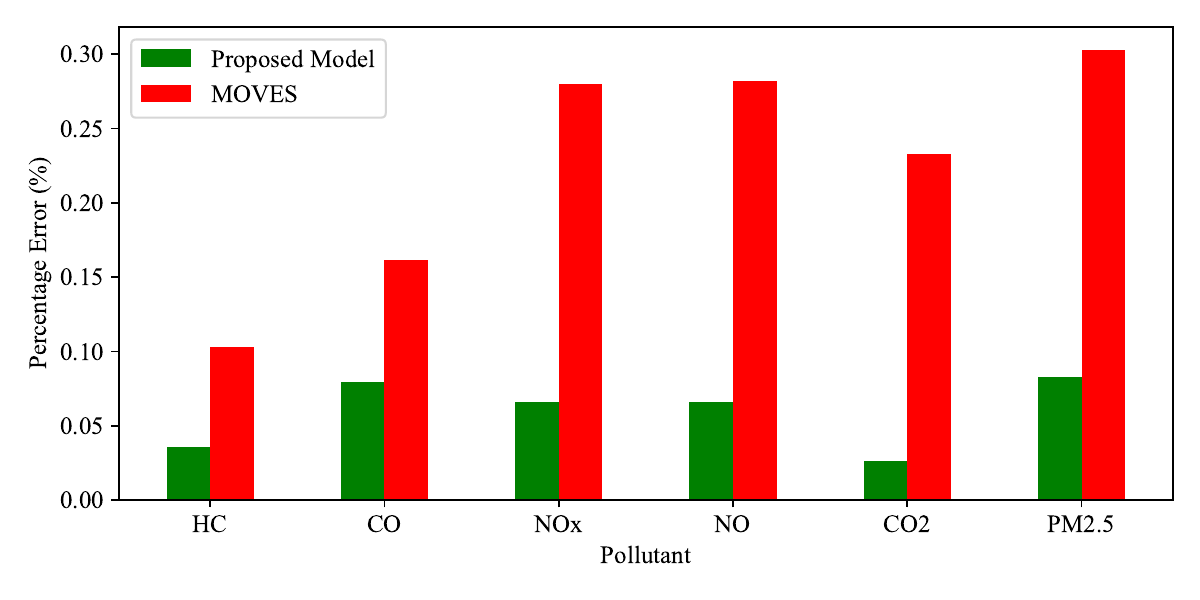}
    \caption{Percentage error in pollutant estimation using the proposed model and MOVES.}
    \label{fig:figure10}
\end{figure}

The operating mode distributions obtained from the two approaches are passed to the relevant module of MOVES to estimate the corresponding emissions. Figure~\ref{fig:figure9} provides a comparison of emissions of different pollutants using the detailed trajectories (used as ground truth), the proposed model, and MOVES. The pollutants analyzed include HC, CO, NOx, NO, CO$_2$, and PM2.5. Each subplot in Figure 8 shows the emissions for a specific pollutant, with the bars representing the actual emissions in blue, the emissions estimated by the proposed model in green, and the emissions estimated by MOVES in red.

For hydrocarbons (HC), the proposed model and MOVES both overestimate emissions compared to the actual data, but the proposed model's estimate is closer to the actual value. In the case of carbon monoxide (CO), the proposed model overestimates emissions, whereas MOVES significantly underestimates them. For nitrogen oxides (NOx) and nitric oxide (NO), both models overestimate emissions, with the proposed model providing a better estimate than MOVES. For carbon dioxide (CO$_2$), both models overestimate ground truth emissions, but the proposed model's estimate is more accurate than MOVES. Lastly, for particulate matter (PM2.5), the proposed model overestimates emissions, whereas MOVES underestimates them, with the proposed model providing a closer estimate to the actual value.

Figure~\ref{fig:figure10} shows the percentage emission estimation error for each pollutant, for the two approaches. The green and red bars represent the percentage error for the proposed model and MOVES, respectively. The results show that for all pollutants, the proposed model exhibits a lower percentage error compared to MOVES. For instance, the proposed model shows significantly lower errors for pollutants such as HC, NOx, and PM2.5. MOVES, on the other hand, consistently has higher percentage errors across the pollutants. The results indicate that emissions estimates based on the default operating modes may be erroneous, and in some cases underestimate important pollutants such as PM2.5.

\section{Conclusion}
The study addresses the problem of developing representative driving cycles that reflect real-world driving behaviors. The study introduces a modular neural network-based approach that aims to estimate the operating mode distributions for a city-wide urban network as a function of easily accessible traffic and network features such as average speed, average volume, number of lanes and traffic control, etc. The study involved a comprehensive microsimulation of an urban traffic network using OD flow data, calibrated against sensor count data. The detailed traffic simulation framework underpinned the development of the NN-based model, which uses macroscopic traffic variables and link infrastructure features to learn the operating mode distributions.

The results from the proposed model show improvements over the traditional MOVES approach to approximate operating mode distribution from just average speed. For instance, the RMSE and R² score highlighted the better performance of the proposed model in almost all operating mode bins. Notably, operating mode bin B1, which represents stop-and-go traffic in urban peak hour conditions, showed the best performance. It underscores the capability of the proposed model to handle complex urban traffic patterns effectively using easily available macroscopic traffic variables, traffic control devices and infrastructure features. Moreover, several bins, such as B13, B21, B24, and B30, exhibited poor performance for MOVES, indicating its limitations, where default driving cycles fail to accurately represent specific local driving conditions. In contrast, the proposed model showed a consistently better performance.

The analysis of emissions estimation validates the performance of the proposed approach, which provided closer estimates to the actual emissions across pollutants like HC, CO, NO, NO$_2$, CO$_2$, and PM2.5. The proposed model achieves an average RMSE of 0.04 in predicting operating mode distributions, compared to 0.08 for MOVES. Furthermore, the average error in emission estimation across pollutants is 8.57\% for the proposed method, lower than the 32.86\% error for MOVES. Notably, for CO$_2$ estimation, the proposed method has an error of just 4\%, compared to 35\% for MOVES.

While the proposed model demonstrates improvements over MOVE, it primarily serves as a proof of concept for the feasibility of this approach. Important future work includes investigating the question of whether the model can be effectively transferred to new areas without requiring re-training. Future research should explore this aspect by conducting similar studies in different regions to assess the consistency and adaptability of the model. The test and training loss curves suggest that the model could be further refined to better fit the actual operating mode distribution. Future work could focus on improving the neural network architecture and fine-tuning the model parameters to enhance its performance.

\bibliographystyle{unsrt}  


\bibliography{references}


\end{document}